\documentclass{article}
\usepackage{iclr2027_conference,times}
\iclrfinalcopy

\usepackage[utf8]{inputenc}
\usepackage[T1]{fontenc}
\usepackage{url}
\usepackage{booktabs}
\usepackage{array}
\usepackage{amsfonts}
\usepackage{amssymb}
\usepackage{nicefrac}
\usepackage{microtype}
\usepackage{xcolor}
\usepackage{graphicx}
\usepackage{algorithm}
\usepackage{algpseudocode}
\usepackage{color,soul}
\usepackage{amsmath}
\usepackage{multirow}
\usepackage{pgfplots}
\usepackage{wrapfig}
\usepackage{needspace}
\usepackage{enumitem}
\pgfplotsset{compat=1.18}
\usepgfplotslibrary{statistics}
\usepackage[hidelinks]{hyperref}
\hypersetup{
  pdftitle={GazeLens for Efficient Egocentric Video Question Answering with Simple Gaze-Driven Adaptive Token Reduction},
  pdfauthor={Trong-Thang Pham, Anh Nguyen, Hien Van Nguyen, Ngan Le}
}

\title{GazeLens for Efficient Egocentric Video\\
Question Answering with Simple Gaze-Driven\\
Adaptive Token Reduction}

\author{Trong-Thang Pham, Anh Nguyen, Hien Van Nguyen, Ngan Le}

\begin{document}

\maketitle
\fancyhead{}
\renewcommand{\headrulewidth}{0pt}

\begin{abstract}
Smart glasses can help users understand their activities and surroundings, but small vision-language models must preserve useful evidence within limited computation. This is difficult because redundant frames consume the visual budget, while aggressive spatial compression can discard relevant details. Although gaze identifies the wearer's focus, the question may require evidence from the surrounding scene. To address these constraints, we propose GazeLens, which combines frame selection with gaze- and question-guided token reduction. First, HOG transitions and image sharpness select frames before visual encoding. For each selected frame, Gaussian pooling then forms gaze-focused and complementary summaries. Because their relevance depends on the question, a lightweight gate blends them into a primary token. Together with a global scene token, this retains both focused and contextual summaries within two tokens per frame. On StreamGaze, GazeLens achieves higher macro and micro accuracy than the evaluated full-grid and token-reduction baselines, with lower input-forward FLOPs than the shared-backbone token-reduction baselines.
\end{abstract}

\section{Introduction}
\label{sec:intro}
\begin{figure}[h!]
    \centering
    \includegraphics[width=0.95\textwidth]{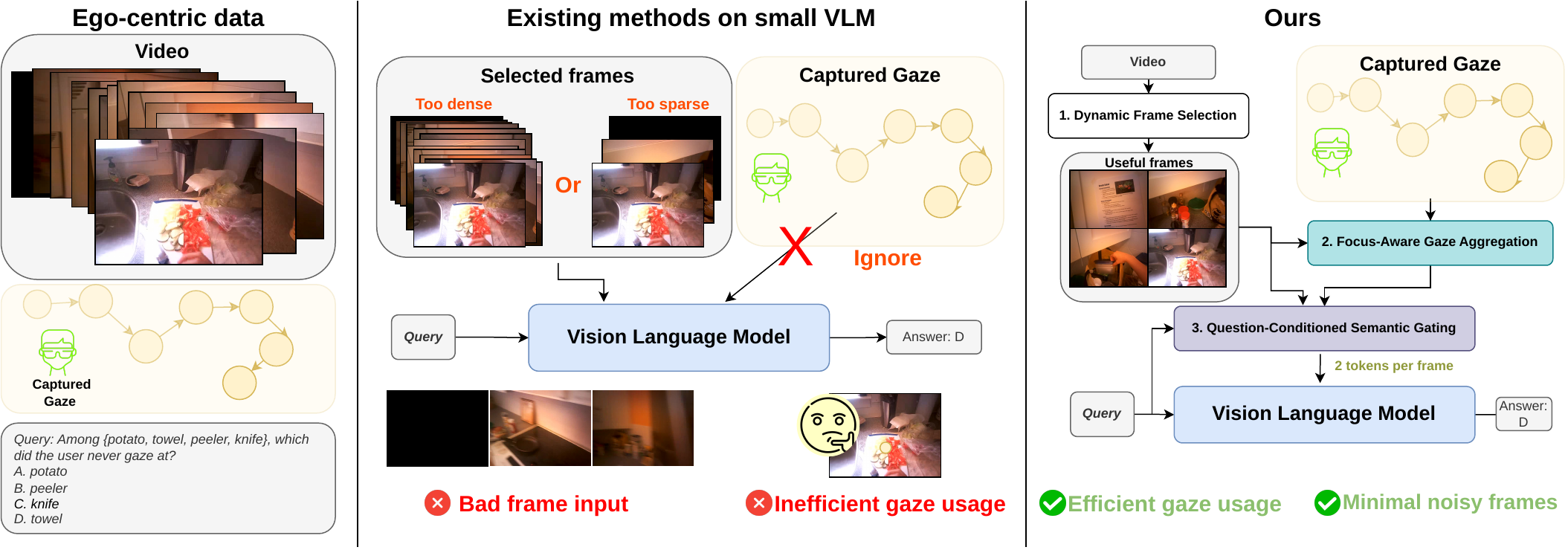}
\caption{
\textbf{Teaser of our gaze-driven adaptive token reduction framework for egocentric video QA.}
\textit{In the left panel,} egocentric video and gaze data are inherently noisy and redundant, with dense frames and unstable gaze signals.
\textit{In the middle panel,} existing VLM approaches suffer from inefficient frame selection (too dense or too sparse) and often ignore gaze or rely on computationally expensive adaptive attention mechanisms.
\textit{In the right panel,} our approach combines
(1) \textbf{Dynamic Frame Selection}, 
(2) \textbf{Focus-Aware Gaze Aggregation}, and 
(3) \textbf{Question-Conditioned Semantic Gating}.
Together, these modules produce two visual tokens per frame. This design reduces the visual token budget and model computation while improving performance under strict token budgets.
}
    \label{fig:teaser}
\end{figure}
Smart glasses can help a wearer recall earlier actions, identify objects, or receive alerts about unfolding events~\citep{streamgaze2025}. However, small vision-language models must retain useful evidence within limited compute and memory~\citep{yao2024minicpm,chu2024mobilevlm}. Egocentric video makes this difficult because repeated views consume the visual budget, while useful evidence may occupy only a small region or a brief interval.

Reducing visual tokens addresses only part of this cost. FastV and PyramidDrop discard tokens inside the language model~\citep{chen2024fastv,xing2024pyramiddrop}, after visual encoding and some language-model computation have already occurred. LLaMA-VID compresses each encoded frame to two tokens~\citep{li2024llamavid}, but still encodes every sampled frame. LongVU reduces both temporal and spatial redundancy~\citep{shen2024longvu}, using DINOv2 features computed before frame removal. These choices save downstream computation while retaining the cost of extracting the features used for reduction. This motivates selecting frames before learned visual encoding when inexpensive image statistics can identify redundancy.

Deciding what to preserve within a frame introduces a second difficulty. Gaze provides a spatial prior for the wearer's focus~\citep{egtea,gazean1,gazean2}. However, a question about a fixated object and a question about the surrounding scene require different evidence. A compact representation must therefore preserve both local and contextual information, with a balance that depends on the question.

Table~\ref{tab:conceptual} compares how reduction mechanisms address these requirements. LLaMA-VID conditions aggregation on the question without a gaze prior. GazeLLM crops around gaze~\citep{rekimoto2025gazellmmultimodalllmsincorporating}, removing off-gaze evidence. GazeVLM preserves global and local image views, then selects gaze-near and surrounding tokens with a preset allocation~\citep{yoon2025gazevlm}. However, this allocation does not depend on the question. Voila-A retains gaze-conditioned features through a learned Perceiver~\citep{voila2024}, adding attention operations after encoding without reducing encoder inputs. These distinctions motivate combining frame selection before encoding with question-conditioned spatial compression that preserves scene context.

GazeLens implements these choices through three modules (Figure~\ref{fig:teaser}). To avoid encoding redundant or blurred candidates, \textbf{Dynamic Frame Selection} uses HOG transitions and Laplacian sharpness to select frames without learned parameters. To keep evidence beyond the fixation available, \textbf{Focus-Aware Gaze Aggregation} forms soft gaze-focused and complementary summaries. \textbf{Question-Conditioned Semantic Gating} then controls their mixture according to the question. This mixture forms the primary token, while a second, global token retains a scene summary regardless of the gate. Together, the modules reduce how many frames are encoded and control which information survives within two tokens per selected frame.

We make three contributions.
\begin{itemize}[nosep]
    \item A parameter-free frame selector reduces encoder inputs using HOG transitions and image sharpness.
    \item Gaze- and question-conditioned pooling preserves focused, complementary, and global summaries within two visual tokens per selected frame.
    \item Our 500M configuration improves StreamGaze Macro and Micro accuracy over the evaluated baselines while reducing model-forward FLOPs. EgoGazeVQA transfer retains the frame-efficiency benefit.
\end{itemize}

\begin{table}[t]
\centering\small
\caption{Coverage of the design requirements. Adaptive frames denotes content-dependent selection before learned visual encoding. The other columns concern spatial compression. A check marks explicit support within the reduction mechanism, including whether compression is conditioned on the question.}
\label{tab:conceptual}
\begin{tabular}{@{}>{\raggedright\arraybackslash}p{0.28\linewidth}*{4}{>{\centering\arraybackslash}p{\dimexpr0.18\linewidth-2\tabcolsep\relax}}@{}}
\toprule
Method & Adaptive frames & Gaze guidance & Peripheral context & Question guidance \\
\midrule
GazeLLM~\citep{rekimoto2025gazellmmultimodalllmsincorporating} & $\times$ & $\checkmark$ & $\times$ & $\times$ \\
\addlinespace[3pt]
LLaMA-VID~\citep{li2024llamavid} & $\times$ & $\times$ & $\checkmark$ & $\checkmark$ \\
\addlinespace[3pt]
Voila-A~\citep{voila2024} & $\times$ & $\checkmark$ & $\checkmark$ & $\times$ \\
\addlinespace[3pt]
GazeVLM~\citep{yoon2025gazevlm} & $\times$ & $\checkmark$ & $\checkmark$ & $\times$ \\
\midrule
\textbf{GazeLens (Ours)} & $\checkmark$ & $\checkmark$ & $\checkmark$ & $\checkmark$ \\
\bottomrule
\end{tabular}
\end{table}

\section{Related works}
\label{sec:related}

\paragraph{Gaze-guided VLMs.} GazeLLM uses gaze-centered cropping~\citep{rekimoto2025gazellmmultimodalllmsincorporating}, while Voila-A learns gaze-conditioned visual representations~\citep{voila2024}. GazeVLM combines global/local image preprocessing with gaze-guided token selection after encoding~\citep{yoon2025gazevlm}. It preserves surrounding evidence through sampled context tokens, but its image-based reduction does not select video frames or condition token allocation on the question. GazeLens instead selects frames before encoding and blends focused and complementary features according to the question within two tokens per frame. StreamGaze~\citep{streamgaze2025} evaluates this balance through Past, Present, and Proactive questions.

\paragraph{Token reduction and frame selection.} FastV and PyramidDrop prune visual tokens within the language model~\citep{chen2024fastv,xing2024pyramiddrop}, whereas LLaMA-VID produces a compact representation before language-model input~\citep{li2024llamavid}. For video, LongVU and FrameFusion also exploit temporal redundancy in visual features~\citep{shen2024longvu,fu2024framefusion}. Frame-Voyager and M-LLM Frame Selector learn to choose frames using question relevance and model-derived supervision~\citep{yu2024framevoyager,hu2025mllmselector}. Across these approaches, the stage of reduction determines which computation can be avoided. We therefore place frame selection before learned encoding and use gaze to allocate the retained spatial-token budget.

\paragraph{Streaming MLLMs.} VideoLLM-online, ViSpeak, Dispider, Flash-VStream, and LiveVLM address interaction with arriving video~\citep{chen2024videollm,fu2025vispeak,qian2025dispider,zhang2024flashvstream,ning2025livevlm}, with StreamingBench evaluating streaming comprehension~\citep{lin2024streamingbench}. Our study concerns the visual input budget for each QA example. Here, frame selection controls the amount of encoded video, and gaze-conditioned pooling controls its spatial representation. Selected frames are encoded independently.

\section{Method}
\label{sec:method}

\begin{figure}[h!]
    \centering
    \includegraphics[width=\textwidth]{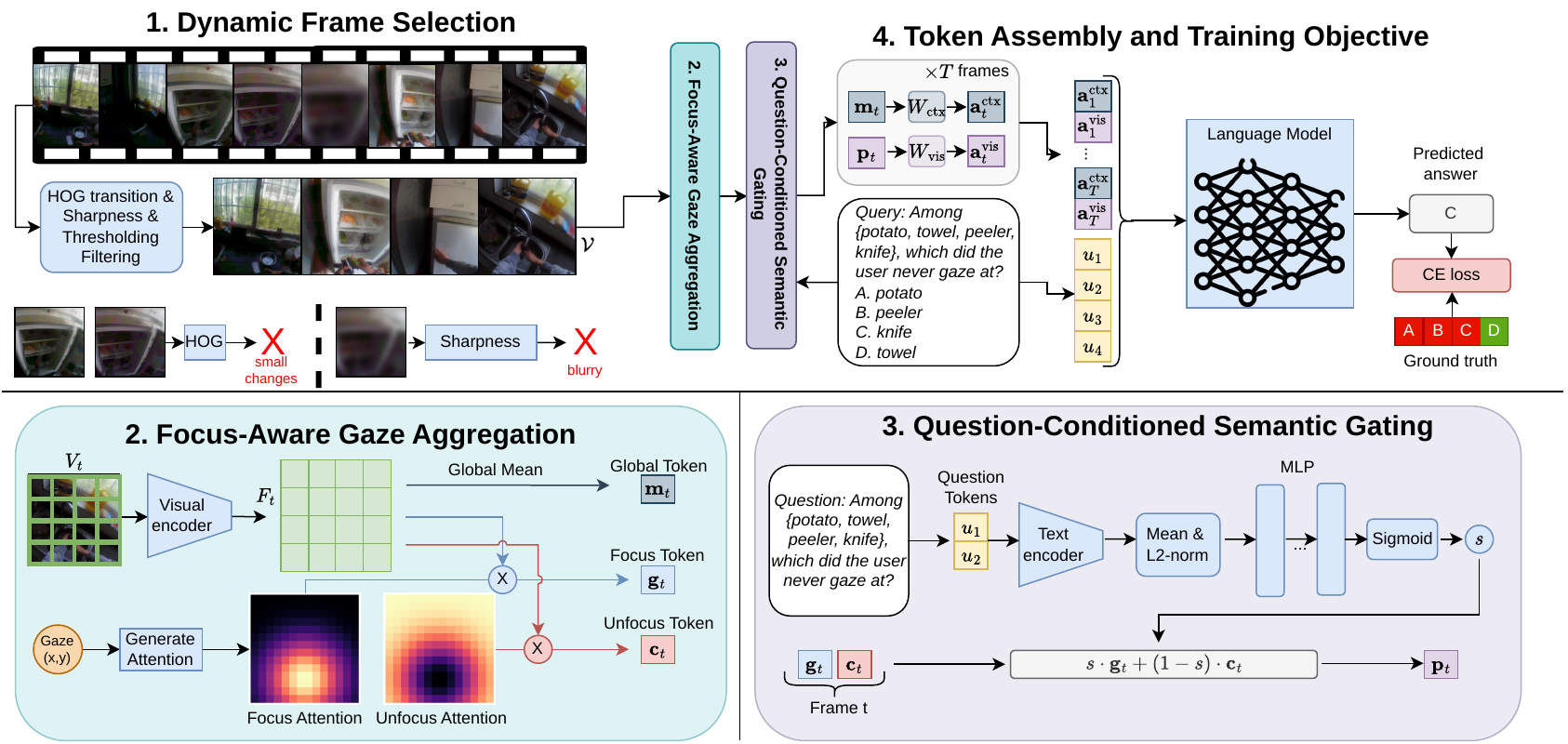}
    \caption{\textbf{Overview of the proposed focus-aware, question-conditioned token reduction framework.}
\textit{(1) Dynamic Frame Selection.} A continuous video stream is filtered using HOG transition and sharpness cues to retain informative frames while discarding small-change and blurry frames. This filtering process produces a dynamic input sequence $\mathcal{V}$.
\textit{(2) Focus-Aware Gaze Aggregation.} Each selected frame is encoded into spatial features and combined with gaze $(x_t, y_t)$ to produce a focus token $\mathbf{g}_t$, an unfocus token $\mathbf{c}_t$, and a global token $\mathbf{m}_t$.
\textit{(3) Question-Conditioned Semantic Gating.} Question token embeddings are mean-pooled and L2-normalized to $\mathbf{t}$. A two-layer MLP and sigmoid then yield a scalar gate $s \in [0,1]$. The gate modulates gaze-focused versus contextual information as $\mathbf{p}_t = s \cdot \mathbf{g}_t + (1 - s) \cdot \mathbf{c}_t$.
\textit{(4) Token Assembly and Training Objective.} The per-frame tokens $\mathbf{p}_t$ and $\mathbf{m}_t$ are linearly projected, concatenated with the question, and fed into a language model to predict the answer, optimized with a cross-entropy (CE) loss.}
    \label{fig:overview}
\end{figure}
Each module in Figure~\ref{fig:overview} is motivated by a distinct failure mode of prior work. Uniform temporal sampling wastes the token budget on redundant or blurry frames before they ever reach the encoder, so Section~\ref{sec:method-dynamic-frame} filters the candidate pool first. Hard gaze crops are brittle to sensor noise and silently drop peripheral context, so Section~\ref{sec:method-gaze} instead builds complementary soft focus and unfocus tokens via Gaussian pooling. Because the right balance between gaze focus and peripheral evidence varies with the question, Section~\ref{sec:method-text} introduces a scalar gate conditioned on question semantics to blend these tokens per sample. Section~\ref{sec:method-vlm} details the final token assembly and training objective.

\subsection{Dynamic Frame Selection}
\label{sec:method-dynamic-frame}

A fixed frame count imposes a single global trade-off. Too few frames miss rare action moments, and too many inflate the LM-side token budget with redundant content. We instead use a variable count $t_{\min}{\le}T{\le}t_{\max}$, spending the budget only on frames where visual content actually changes. This flexibility favors frames that are both visually distinct from their neighbors and free of motion blur. We operationalize these two criteria with HOG transition cues and Laplacian sharpness scores, which are parameter-free, fast to compute, and require no training data. 

\paragraph{HOG transition cue.}
Sharpness alone cannot distinguish a novel sharp frame from a sharp duplicate, so HOG transition cues are added to suppress consecutive frames with near-identical visual content.
Let $(I_{i-1},I_i)$ denote consecutive frames after striding.
Both are resized to a fixed HOG window $H_r{\times}W_r$ and converted to grayscale.
Let $\phi(I_i)\in\mathbb{R}^d$ be the single-window HOG descriptor of $I_i$.
We record the $L_2$ distance between consecutive descriptors at the later index $i$,
\begin{equation}
\delta_i \;=\; \bigl\lVert \phi(I_i)-\phi(I_{i-1})\bigr\rVert_2.
\label{eq:hog-dist}
\end{equation}
Large $\delta_i$ indicate strong layout or motion change between successive kept frames. HOG encodes gradient orientations, making $\delta_i$ robust to the illumination shifts and auto-exposure changes common in egocentric footage.

\paragraph{Sharpness cue.}
Independently, each frame is resized to $H_r{\times}W_r$ grayscale and scored by the variance of its $3{\times}3$ Laplacian,
\begin{equation}
\ell_i \;=\; \mathrm{Var}\!\bigl(\mathrm{Lap}_{3\times 3}(I_i)\bigr).
\label{eq:sharp}
\end{equation}
Higher $\ell_i$ correspond to sharper, less motion-blurred content. Motion blur suppresses high-frequency edges and collapses the Laplacian variance to near zero, giving a calibration-free score that drops precisely on frames with no recoverable spatial detail.

\paragraph{Scoring and selection.}
Transition cues alone tend to peak on the blurred interstitial frame between two sharp moments, while sharpness scores alone favour static frames that carry no new information, so combining them selects frames that are both informationally novel and spatially recoverable.
For each cue $c\in\{\delta,\ell\}$, let $\Omega_c=\{i: c_i\}$, $N_c=|\Omega_c|$, and define the per-clip mean
$\mu_c = \frac{1}{N_c}\sum_{i\in\Omega_c} c_i$, and variance 
$\sigma_c^2 = \frac{1}{N_c}\sum_{i\in\Omega_c}(c_i-\mu_c)^2$.
Per-clip normalized scores are $\tilde{c}_i =
\dfrac{c_i-\mu_c}{\sigma_c}.$
We combine the two cues with fixed weights $w_{\mathrm{hog}}$ and $w_{\mathrm{sh}}$ to obtain
\begin{equation}
S_i \;=\; \sum_{\substack{c\in\{\delta,\ell\}}} w_c\,\tilde{c}_i.
\label{eq:final_score}
\end{equation}

We retain frame candidates with $S_i > z_{\mathrm{thresh}}$ and sort them by $S_i$.
If more than $t_{\max}$ indices remain, we truncate the set to the top-$S_i$ entries to enforce a hard cap on encoder cost and the downstream token budget.

Each retained index is then replaced by $\arg\max_{j:|j-i|\le \tau} \ell_j$, where $\tau$ denotes the scan window. This adjustment is motivated by the observation that pairwise change cues often peak on motion-blurred transition frames, whereas larger $\ell_j$ values typically occur on nearby stable frames.

If fewer than $t_{\min}$ indices remain, e.g., in nearly static clips, we supplement the set with uniformly spaced frames to ensure the VLM still receives a minimum level of temporal information.

Both $z_{\mathrm{thresh}}$ and $\tau$ are hyperparameters selected based on training set analysis.
The selected frames define our dynamic input sequence $\mathcal{V} = \{V_t\}_{t=0}^{T-1}$.
Both cues are parameter-free image statistics.

Temporal reduction occurs before visual encoding. Each selected frame is then encoded independently, and tokens are not reused between frames.

\subsection{Focus-Aware Gaze Aggregation}
\label{sec:method-gaze}

For the selected $T$ frames, let $\mathcal{G} = \{(x_t, y_t)\}_{t=0}^{T-1}$ denote the gaze points normalized to $[0,1]^2$. We then encode each frame $t$ to obtain spatial patch features $F_t \in \mathbb{R}^{H \times W \times D}$, which provide the input to gaze-guided pooling.

Hard-cropping at the raw gaze point $(x_t, y_t)$ is brittle, as sensor jitter of a few pixels selects an entirely different patch and the crop discards all peripheral context. Instead, we generate an isotropic Gaussian map over the $H \times W$ feature grid, which smoothly down-weights patches away from the gaze point, preserving peripheral context with graduated weights.
\begin{equation}
(\widetilde{h}_t)_{i,j} = \exp\!\left(-\frac{(i - y_t (H{-}1))^2 + (j - x_t (W{-}1))^2}{2 \rho^2}\right), \qquad \rho = \max\!\left(\tfrac{\max(H, W)}{4},\, 0.5\right).
\label{eq:gauss}
\end{equation}
Subsequently, we normalize the Focus and Unfocus branches using $h_t = \widetilde{h}_t / \sum \widetilde{h}_t$ for Focus and $\bar{h}_t = (\mathbf{1} - \widetilde{h}_t) / \sum (\mathbf{1} - \widetilde{h}_t)$ for Unfocus. This yields the two pooled vectors on the patch features
\begin{equation}
\mathbf{g}_t \;=\; \sum_{i,j} (h_t)_{i,j}\, (F_t)_{i,j} \;\in\; \mathbb{R}^D, \qquad \mathbf{c}_t \;=\; \sum_{i,j} (\bar{h}_t)_{i,j}\, (F_t)_{i,j} \;\in\; \mathbb{R}^D.
\label{eq:pool}
\end{equation}
Here, $\mathbf{g}_t$ aggregates gaze-focused evidence, whereas $\mathbf{c}_t$ emphasizes its complement. Thus, peripheral evidence remains available to the question gate through $\mathbf{c}_t$.

In addition, we compute a Global token $\mathbf{m}_t = \tfrac{1}{HW} \sum_{i,j} (F_t)_{i,j}$, which provides a gaze-independent scene summary. If gaze coordinates are absent or erratic for a frame, the LM retains an unbiased holistic representation without depending on noisy $(x_t, y_t)$ values.

\subsection{Question-Conditioned Semantic Gating}
\label{sec:method-text}
Although the gaze heatmap highlights the user's focus, some questions also require peripheral evidence. We therefore condition a semantic gate on the question to control the balance between these two sources.

We represent the corresponding question as $\mathbf{u} \in \mathbb{Z}^{L_u}$. The question $\mathbf{u}$ is tokenized and embedded using the VLM's text encoder $E$. Mean-pooling collapses variable-length questions into a fixed-size vector without additional parameters, and L2-normalization removes the scale difference between short and long questions so gate outputs are comparable across samples.
\begin{equation}
\bar{\mathbf{e}} \;=\; \frac{1}{L_u}\sum_{k=1}^{L_u} E(u_k),\qquad \mathbf{t} \;=\; \frac{\bar{\mathbf{e}}}{\lVert \bar{\mathbf{e}} \rVert_2 + \varepsilon} \;\in\; \mathbb{R}^D,\qquad \varepsilon = 10^{-6}.
\end{equation}
We then pass the normalized vector $\mathbf{t}$ through a lightweight two-layer Multi-Layer Perceptron (MLP), $\phi_\theta: \mathbb{R}^D \to \mathbb{R}$, followed by a sigmoid to obtain the gating scalar
\begin{equation}
s \;=\; \sigma\!\big(\phi_\theta(\mathbf{t})\big) \;\in\; [0, 1].
\end{equation}
A scalar directly controls the balance between focused and complementary evidence. Because it encodes a preference from the question, we share this coefficient across the clip's frames. We then form the primary contextual token by blending the two pooled representations.
\begin{equation}
\mathbf{p}_t \;=\; s \cdot \mathbf{g}_t \;+\; (1 - s) \cdot \mathbf{c}_t.
\label{eq:gate}
\end{equation}
When the gate $s \to 1$, the primary token concentrates on the gaze region. When $s \to 0$, it is drawn from the complementary visual evidence.

\subsection{Token Assembly and Training Objective}
\label{sec:method-vlm}

To retain both question-specific and scene-level evidence, we pair the gated token $\mathbf{p}_t$ with the global token $\mathbf{m}_t$. The former carries the question-conditioned mixture of gaze and complementary features, while the latter provides a scene summary independent of gaze. We then map both tokens into the language-model representation using trainable linear projectors.
\begin{equation}
\mathbf{a}^{\text{ctx}}_t = W_{\text{ctx}}\, \mathbf{p}_t,\qquad \mathbf{a}^{\text{vis}}_t = W_{\text{vis}}\, \mathbf{m}_t,\qquad W_{\text{ctx}}, W_{\text{vis}} \in \mathbb{R}^{D \times D}.
\end{equation}
The ordering for each frame is established as $[\mathbf{a}^{\text{ctx}}_t, \mathbf{a}^{\text{vis}}_t]$, acting as a visual prefix for the text. Thus, the total visual budget on the LM side is strictly limited to $2T$.

We frame the egocentric video QA task as a multiple-choice question answering (MCQA) problem. During training, we keep the backbone VLM base weights frozen and attach lightweight LoRA adapters to the attention layers, minimizing the trainable parameter footprint while preserving pre-trained reasoning capabilities. Gradients update only the LoRA adapters, the linear projectors ($W_{\text{ctx}}$, $W_{\text{vis}}$), and the semantic gate $\phi_\theta$.

The entire framework is optimized end-to-end using a Cross-Entropy (CE) loss over the probability distribution of the four answer options (A, B, C, D) with label smoothing.
\begin{equation}
\mathcal{L} = - \sum_{k \in \{A, B, C, D\}} \left( (1 - \alpha) y_k + \frac{\alpha}{4} \right) \log P(\text{token}_k \mid \mathcal{V}, \mathcal{G}, \mathbf{u}),
\end{equation}
where $y_k \in \{0, 1\}$ is the ground-truth binary label, $\alpha$ is the label smoothing factor, and $P(\text{token}_k \mid \mathcal{V}, \mathcal{G}, \mathbf{u})$ is the predicted probability generated by the LM at the final sequence position.

\section{Experiments}
\label{sec:exp}

We first compare StreamGaze accuracy and model-forward computation, then assess transfer to EgoGazeVQA. Finally, component interventions examine sensitivity to frame selection and spatial pooling. Appendix~\ref{sec:supp-robustness} reports gaze-noise controls.

\subsection{Implementation}
\label{sec:exp-implementation}

\textbf{Datasets \& Metrics.} We evaluate our method on StreamGaze~\citep{streamgaze2025}, an egocentric video QA benchmark that we evaluate as three tasks, Past, Present, and Proactive, covering ten subtasks drawn from EGTEA Gaze+~\citep{egtea}, EgoExo~\citep{grauman2024ego}, and HoloAssist~\citep{wang2023holoassist}. Following the official partition, we train on 5,766 QA pairs and report final results on the held-out test split of 1,399 questions, yielding 2,418 evaluation examples after expanding proactive questions into timestamped probes. We adopt macro accuracy and micro accuracy as the primary evaluation metrics. Each task score averages its constituent subtask accuracies, and Macro is the equal-weight mean of the Past, Present, and Proactive scores. Micro is accuracy pooled over all test examples. To quantify model computation, we report TFLOPs per input forward under the controlled protocol in Appendix~\ref{sec:controlled-efficiency}.

\textbf{Baselines \& Models.} We compare our method against two groups of baselines. The first group consists of small VLMs evaluated without token reduction or gaze tuning, including SmolVLM2-256M and SmolVLM2-500M~\citep{marafioti2025smolvlm} (our 256M and 500M backbones), FastVLM~\citep{fastvlm2025}, and Mobile-VideoGPT~\citep{Shaker2025MobileVideoGPT}. These represent the accuracy achievable with full token grids and standard inference pipelines, none of which were evaluated in the original StreamGaze report. The second group consists of token-reduction methods re-implemented on the SmolVLM2-500M framework for a controlled comparison of visual token budget and accuracy on identical hardware, namely GazeLLM~\citep{rekimoto2025gazellmmultimodalllmsincorporating}, LLaMA-VID~\citep{li2024llamavid}, and Voila-A~\citep{voila2024}. We use SmolVLM2-500M as the shared backbone to control for model capacity. We select these three because they each reduce visual tokens through distinct mechanisms. GazeLLM crops each frame to a fixed patch centered on the raw gaze point, discarding all peripheral context. LLaMA-VID produces two tokens per frame by combining one global mean over the feature grid with one context token formed by text-attended averaging over the full feature grid without any gaze prior. Voila-A feeds a gaze heatmap into a Perceiver resampler and mixes the resampled output with intermediate visual features.

\textbf{Hyperparameters.} Videos are first subsampled to 1 FPS before frame selection, normalizing the candidate pool across clips that vary widely in native frame rate. The dynamic frame selector then operates in the range $t_{min}=4$ to $t_{max}=16$ frames per clip, with per-video scores cached to avoid redundant computation across multiple questions from the same video. Our framework strictly limits the visual span to 2 tokens per frame. The semantic gate is a two-layer MLP with ReLU activation and sigmoid output. The gaze heatmap uses $\rho = \max(\max(H, W)/d,\ 0.5)$ with $d=4$. We train the semantic gate, linear projectors ($W_{\text{ctx}}$, $W_{\text{vis}}$), and LoRA~\citep{hu2022lora} adapters (rank $r=8$, $\alpha=16$, targeting $q_{\text{proj}}$ and $v_{\text{proj}}$) using AdamW with cross-entropy loss and label smoothing ($\varepsilon=0.1$). Training runs for 4 epochs on an NVIDIA A6000 GPU.

\subsection{Results}
\label{sec:exp-main}
\begin{table}[t]
\centering
\caption{StreamGaze accuracy and model cost, grouped by total parameter count. Params counts unique parameters, including vision encoders and added modules. TFLOPs are per input forward (Appendix~\ref{sec:controlled-efficiency}). Macro equally averages Past, Present, and Proactive, while Micro pools test examples. Both Ours variants report three-seed means. Across-seed variability is reported separately in Table~\ref{tab:seed-variability}. Bold/italic mark best/second-best accuracy. Adap. denotes dynamic selection, SVLM denotes SmolVLM2, and Mob-VGPT denotes Mobile-VideoGPT.}
\label{tab:group-results}
\resizebox{\textwidth}{!}{\begin{tabular}{lrrr ccccc}
\toprule
Method & Params (M) $\downarrow$ & TFLOPs $\downarrow$ & Fr. & Past $\uparrow$ & Present $\uparrow$ & Proactive $\uparrow$ & Macro $\uparrow$ & Micro $\uparrow$ \\
\midrule
\multicolumn{9}{l}{\emph{Below 300M parameters}} \\
SVLM-256M & 256.48 & 3.966 & 16 & 0.328 & 0.283 & 0.537 & 0.383 & 0.480 \\
\textbf{Ours-256M} & 258.24 & 2.857 & Adap. & 0.351 & \textit{0.532} & 0.643 & \textit{0.509} & \textit{0.617} \\
\midrule
\multicolumn{9}{l}{\emph{300M+ parameters}} \\
SVLM-500M & 507.48 & 4.742 & 16 & 0.350 & 0.449 & 0.578 & 0.459 & 0.582 \\
GazeLLM & 508.60 & 3.791 & 16 & 0.309 & 0.233 & 0.636 & 0.393 & 0.502 \\
LLaMA-VID & 510.44 & 3.804 & 16 & \textit{0.378} & 0.436 & 0.635 & 0.483 & 0.576 \\
\textbf{Ours-500M} & 511.37 & 3.055 & Adap. & \textbf{0.423} & \textbf{0.551} & \textit{0.645} & \textbf{0.539} & \textbf{0.640} \\
Voila-A & 530.39 & 4.758 & 16 & 0.311 & 0.503 & \textbf{0.650} & 0.488 & 0.610 \\
FastVLM & 622.66 & 13.591 & 16 & 0.377 & 0.412 & 0.476 & 0.421 & 0.466 \\
Mob-VGPT & 656.36 & 3.903 & 16 & 0.350 & 0.370 & 0.507 & 0.409 & 0.492 \\
\bottomrule
\end{tabular}}
\end{table}

Table~\ref{tab:group-results} compares accuracy, total parameters, and counted input-forward computation. Across three seeds, our 256M configuration achieves $50.9\%$ Macro and $61.66\pm0.50\%$ Micro. With the larger backbone, our 500M configuration achieves $53.94\pm1.54\%$ macro and $64.01\pm0.49\%$ micro accuracy (mean $\pm$ sample standard deviation). At the same time, Ours-500M uses 3.055 TFLOPs per input, compared with 3.791 for GazeLLM, 3.804 for LLaMA-VID, and 4.758 for Voila-A under the shared-backbone protocol.

The methods differ in how they allocate the compact visual budget. Soft Gaussian pooling retains peripheral evidence that GazeLLM's crop discards, while the gate controls its contribution to the primary token. LLaMA-VID also uses two tokens per frame, but obtains its context token through text-conditioned attention over the full grid. GazeLens adds the wearer's gaze as a spatial prior and preserves a global summary alongside it.

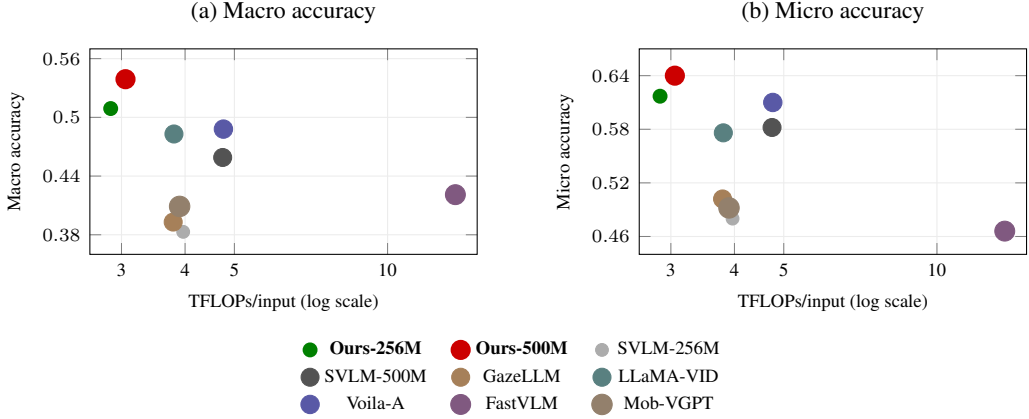
\begin{figure}[t]
\centering
\begin{minipage}{0.48\textwidth}
\centering
\begin{tikzpicture}
\begin{axis}[
width=\linewidth,height=4.3cm,
title={(a) Macro accuracy},title style={font=\small},
xlabel={TFLOPs/input (log scale)},ylabel={Macro accuracy},
xmode=log,log basis x=10,xmin=2.6,xmax=15,xtick={3,4,5,10},xticklabels={3,4,5,10},
ymin=0.36,ymax=0.57,ytick={0.38,0.44,0.50,0.56},
tick label style={font=\scriptsize},label style={font=\scriptsize},
grid=major,grid style={gray!15},
legend to name=efficiency-methods,legend columns=3,legend style={font=\scriptsize,draw=none,/tikz/every even column/.append style={column sep=7pt}},
]
\addplot[only marks,mark=*,mark size=2.408189pt,green!50!black,thick] coordinates {(2.857,0.509)};
\addlegendentry{\textbf{Ours-256M}}
\addplot[only marks,mark=*,mark size=3.388799pt,red!80!black,thick] coordinates {(3.055,0.539)};
\addlegendentry{\textbf{Ours-500M}}
\addplot[only marks,mark=*,mark size=2.400000pt,gray!65] coordinates {(3.966,0.383)};
\addlegendentry{SVLM-256M}
\addplot[only marks,mark=*,mark size=3.375909pt,black!35!gray] coordinates {(4.742,0.459)};
\addlegendentry{SVLM-500M}
\addplot[only marks,mark=*,mark size=3.379613pt,orange!30!gray] coordinates {(3.791,0.393)};
\addlegendentry{GazeLLM}
\addplot[only marks,mark=*,mark size=3.385738pt,teal!30!gray] coordinates {(3.804,0.483)};
\addlegendentry{LLaMA-VID}
\addplot[only marks,mark=*,mark size=3.451251pt,blue!30!gray] coordinates {(4.758,0.488)};
\addlegendentry{Voila-A}
\addplot[only marks,mark=*,mark size=3.739444pt,violet!30!gray] coordinates {(13.591,0.421)};
\addlegendentry{FastVLM}
\addplot[only marks,mark=*,mark size=3.839303pt,brown!30!gray] coordinates {(3.903,0.409)};
\addlegendentry{Mob-VGPT}
\end{axis}
\end{tikzpicture}
\end{minipage}
\hfill
\begin{minipage}{0.48\textwidth}
\centering
\begin{tikzpicture}
\begin{axis}[
width=\linewidth,height=4.3cm,
title={(b) Micro accuracy},title style={font=\small},
xlabel={TFLOPs/input (log scale)},ylabel={Micro accuracy},
xmode=log,log basis x=10,xmin=2.6,xmax=15,xtick={3,4,5,10},xticklabels={3,4,5,10},
ymin=0.44,ymax=0.67,ytick={0.46,0.52,0.58,0.64},
tick label style={font=\scriptsize},label style={font=\scriptsize},
grid=major,grid style={gray!15},
]
\addplot[only marks,mark=*,mark size=2.408189pt,green!50!black,thick] coordinates {(2.857,0.617)};
\addplot[only marks,mark=*,mark size=3.388799pt,red!80!black,thick] coordinates {(3.055,0.640)};
\addplot[only marks,mark=*,mark size=2.400000pt,gray!65] coordinates {(3.966,0.480)};
\addplot[only marks,mark=*,mark size=3.375909pt,black!35!gray] coordinates {(4.742,0.582)};
\addplot[only marks,mark=*,mark size=3.379613pt,orange!30!gray] coordinates {(3.791,0.502)};
\addplot[only marks,mark=*,mark size=3.385738pt,teal!30!gray] coordinates {(3.804,0.576)};
\addplot[only marks,mark=*,mark size=3.451251pt,blue!30!gray] coordinates {(4.758,0.610)};
\addplot[only marks,mark=*,mark size=3.739444pt,violet!30!gray] coordinates {(13.591,0.466)};
\addplot[only marks,mark=*,mark size=3.839303pt,brown!30!gray] coordinates {(3.903,0.492)};
\end{axis}
\end{tikzpicture}
\end{minipage}
\par\smallskip
\pgfplotslegendfromname{efficiency-methods}
\caption{Accuracy versus input-forward computation at each model's native image resolution. Muted colors identify baselines, while green and red highlight Ours-256M and Ours-500M. Linear marker size scales with the square root of total parameters. Higher and further left is better. The horizontal axis is logarithmic. FLOPs exclude frame selection, preprocessing, and answer generation (Appendix~\ref{sec:controlled-efficiency}).}
\label{fig:flops-accuracy}
\end{figure}

However, Voila-A achieves slightly higher Proactive accuracy than Ours-500M (0.650 vs.\ 0.645), with its advantage concentrated in gaze-triggered alerts, whereas ours performs better on object-appearance alerts. This advantage comes with higher counted input-forward computation of 4.758 versus 3.055 TFLOPs per input, about 56\% more under the RTX 3080 profiling protocol (Figure~\ref{fig:flops-accuracy}). Voila-A processes 16 frames with a gaze-conditioned Perceiver, whereas ours averages 12.72 selected frames and uses two visual tokens per frame. The Perceiver may retain spatial cues useful for alert triggering. Temporal sampling and spatial representation both differ between these methods. Overall, ours retains higher Macro and Micro accuracy.

Beyond StreamGaze, Section~\ref{sec:cross-dataset} evaluates transfer to EgoGazeVQA. Appendices~\ref{sec:detailed-ablation}, \ref{sec:additional-results}, and~\ref{sec:gaze-robustness} examine frame selection, spatial representations, and gaze robustness, respectively. Appendix~\ref{sec:controlled-efficiency} reports the controlled input-forward FLOP protocol.

\Needspace{20\baselineskip}
\subsection{Cross-dataset Generalization}
\label{sec:cross-dataset}
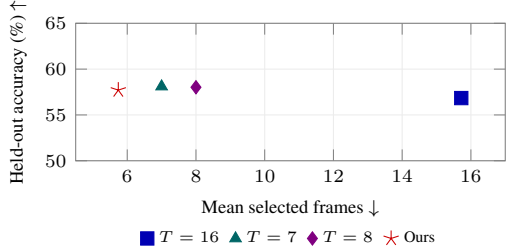
\begin{wrapfigure}{r}{0.52\textwidth}
\centering
\begin{tikzpicture}
\begin{axis}[
width=\linewidth,height=3.4cm,
xlabel={Mean selected frames $\downarrow$},
ylabel={Held-out accuracy (\%) $\uparrow$},
xmin=4.5,xmax=17,ymin=50,ymax=65,
xtick={6,8,10,12,14,16},ytick={50,55,60,65},
tick label style={font=\scriptsize},label style={font=\scriptsize},
grid=major,grid style={gray!15},axis line style={gray},tick style={gray},
legend style={font=\tiny,draw=none,at={(0.5,-0.43)},anchor=north,legend columns=4,/tikz/every even column/.append style={column sep=2pt}},
]
\addplot[only marks,mark=square*,mark size=2.5pt,blue!70!black] coordinates {(15.72,56.84)};
\addlegendentry{$T=16$}
\addplot[only marks,mark=triangle*,mark size=2.5pt,teal!80!black] coordinates {(7.00,58.10)};
\addlegendentry{$T=7$}
\addplot[only marks,mark=diamond*,mark size=2.5pt,violet] coordinates {(8.00,58.02)};
\addlegendentry{$T=8$}
\addplot[only marks,mark=star,mark size=3pt,red!80!black] coordinates {(5.74,57.71)};
\addlegendentry{Ours}
\end{axis}
\end{tikzpicture}
\caption{EgoGazeVQA transfer on 1,265 held-out examples. $T$ denotes uniform sampling, whereas ours is dynamic. Higher and further left is better.}
\label{fig:transfer-efficiency}
\end{wrapfigure}

To assess transfer beyond StreamGaze, we train for four epochs on the EGTEA source of EgoGazeVQA and evaluate on held-out Ego4D and EgoExo examples. Figure~\ref{fig:transfer-efficiency} reports accuracy pooled over evaluation examples. Uniform budgets of seven and eight frames provide comparisons at frame budgets close to the dynamic configuration.

Dynamic selection uses the fewest frames and exceeds Uniform $T=16$ in accuracy, while Uniform $T=7$ and $T=8$ achieve slightly higher accuracy. These results show that adaptive selection retains competitive accuracy with a smaller frame budget on a second gaze-VQA benchmark, supporting the transfer of its frame-efficiency benefit beyond StreamGaze. Appendix~\ref{sec:transfer-details} provides the source-specific breakdown.

\subsection{Ablation Studies}
\label{sec:exp-ablation}

\begin{figure}[t]
    \centering
    \includegraphics[width=\textwidth]{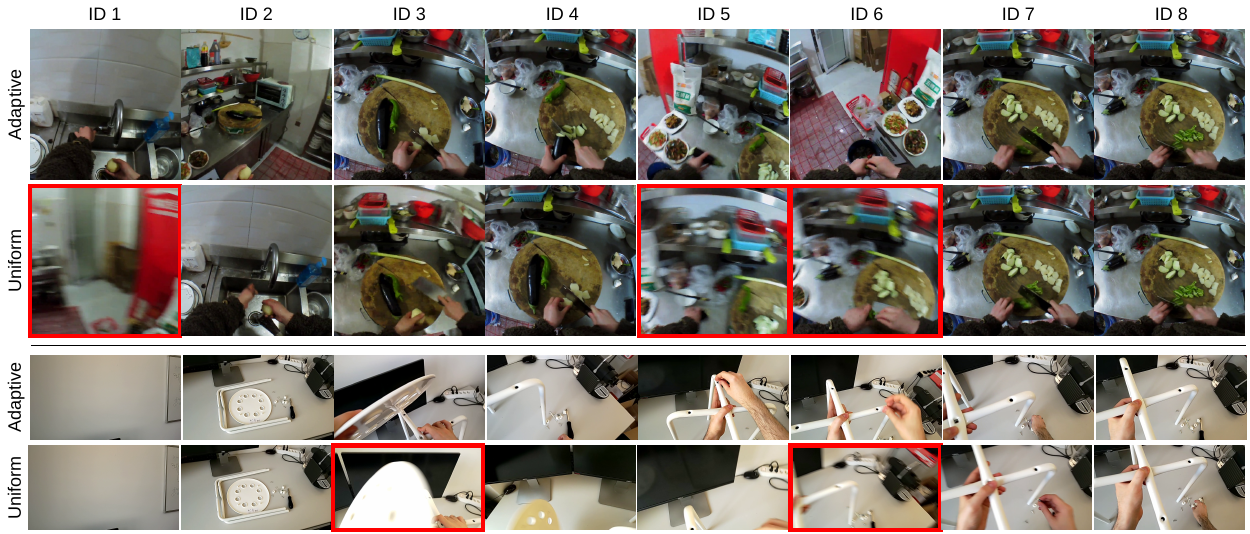}
\caption{Adaptive versus uniform frame selection for example video sequences. Red boxes mark redundant or motion-blurred frames in the uniform selections. Frame IDs run from 1 to 8.}
    \label{fig:dynamic}
\end{figure}
\textbf{Protocol.} To measure sensitivity to frame selection and the mixing coefficient, we change one component at inference while keeping trained weights fixed. The full-method reference uses three Ours-500M checkpoints selected by validation micro accuracy. As in the main comparison, Macro equally averages Past, Present, and Proactive. Mean and sample SD summarize the full-method reference across checkpoints.

\textbf{Frame selection.} The picker selects 12.72 frames per test example on average (SD 3.86 across examples, median 14, range 4--16). Frame counts are identical across the three checkpoints because selection is independent of learned weights. To compare temporal policies at a fixed spatial-token budget, Table~\ref{tab:ablate-uniform-dynamic} replaces the HOG-and-sharpness selector with uniform sampling of up to 16 frames from the same 1-FPS candidate sequence. The dynamic configuration retains between 4 and 16 frames. Both settings use the same gaze pooling, learned gate, and two visual tokens per selected frame. Figure~\ref{fig:dynamic} illustrates the selected frames.

\textbf{Learned gate.} To examine the role of the learned mixture, Table~\ref{tab:ablate-gate-fixed} overrides the learned mixing coefficient with $s=1$, $s=0.5$, or $s=0$ during evaluation. These settings use gaze-focused pooling, an equal mixture, or complementary pooling in the primary token, respectively. The global token and selected frames remain unchanged. Appendix~\ref{sec:detailed-ablation} gives additional component comparisons.

\begin{table}[t]
\centering
\begin{minipage}[t]{0.48\textwidth}
\centering
\caption{Frame-selection ablation. Ours reports mean Macro across three checkpoints, and $\bar T$ is mean frames per evaluation example.}
\label{tab:ablate-uniform-dynamic}
\small
\begin{tabular}{lrr}
\toprule
Selector & $\bar T$ & Macro \\
\midrule
Uniform $T=16$ & 16.00 & 0.525 \\
Dynamic & 12.72 & 0.539 \\
\bottomrule
\end{tabular}
\end{minipage}\hfill
\begin{minipage}[t]{0.48\textwidth}
\centering
\caption{Inference-time gate ablation. Fixed gates replace the learned mixing coefficient.}
\label{tab:ablate-gate-fixed}
\small
\begin{tabular}{lr}
\toprule
Gate & \shortstack{Macro} \\
\midrule
Gaze only ($s=1$) & 0.524 \\
Equal blend ($s=0.5$) & 0.529 \\
Complement only ($s=0$) & 0.520 \\
Learned gate & 0.539 \\
\bottomrule
\end{tabular}
\end{minipage}
\end{table}

\Needspace{17\baselineskip}
\begin{wrapfigure}[11]{r}{0.48\textwidth}
\centering
\begin{tikzpicture}
\begin{axis}[
width=\linewidth,height=3.5cm,
boxplot/draw direction=y,
ylabel={Gate $s$},
xmin=0.4,xmax=3.6,ymin=0.55,ymax=0.72,
xtick={1,2,3},
xticklabels={Past,Present,Proactive},
xticklabel style={font=\scriptsize},yticklabel style={font=\scriptsize},
grid=major,grid style={gray!15},
]
\addplot[boxplot prepared={lower whisker=0.593750000000,lower quartile=0.611328125000,median=0.621093750000,upper quartile=0.652343750000,upper whisker=0.664062500000,draw position=1},draw=blue!65!black,fill=blue!10] coordinates {};
\addplot[boxplot prepared={lower whisker=0.574218750000,lower quartile=0.652343750000,median=0.687500000000,upper quartile=0.687500000000,upper whisker=0.691406250000,draw position=2},draw=blue!65!black,fill=blue!10] coordinates {};
\addplot[boxplot prepared={lower whisker=0.640625000000,lower quartile=0.640625000000,median=0.644531250000,upper quartile=0.644531250000,upper whisker=0.648437500000,draw position=3},draw=blue!65!black,fill=blue!10] coordinates {};
\end{axis}
\end{tikzpicture}
\caption{Gate distributions from one evaluated checkpoint. Boxes show quartiles and medians, while whiskers span min--max. Groups pool evaluation rows, including repeated question templates.}
\label{fig:gate-seed1}
\end{wrapfigure}
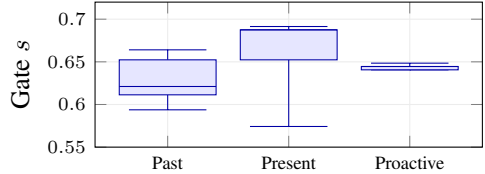
\textbf{Gate behavior.} To examine how the learned gate allocates attention, Figure~\ref{fig:gate-seed1} shows one evaluated Ours-500M checkpoint on the StreamGaze test set. Larger $s$ favors gaze-focused pooling, while the global token remains present. Across examples, the mean is 0.6509 and the within-run standard deviation is 0.0253. Present has the highest median gaze weight. Each value averages visual-block gate outputs, and evaluation rows receive equal weight within each group.

\section{Conclusion}
\label{sec:conclusion}

In this paper, we proposed GazeLens for compact egocentric video question answering. First, adaptive frame selection uses HOG transitions and image sharpness to reduce temporal redundancy before visual encoding. Next, gaze-focused and complementary pooling preserve evidence within and beyond fixation, while a question-conditioned gate adjusts their contribution. A global summary then retains scene context alongside the primary token, yielding two visual tokens per selected frame. Together, these modules address the need to preserve relevant evidence within a limited visual budget. Future work could reuse visual tokens across frames with little change to reduce repeated computation during longer interactions. In addition, adapting the spatial token budget to the question and scene complexity could preserve finer details when needed.

\clearpage
\subsection*{AI Use Statement}

In this work, we used generative AI tools (i.e. ChatGPT) for revising passages, and correcting grammar.
 
All research ideas, methods, experiments, analyses, and conclusions are the authors' own, and the authors reviewed and take full responsibility for all content of this paper.

\subsection*{Ethics Statement}
\label{sec:broader-impact}
This work studies egocentric video and gaze data from existing benchmarks. Such data can expose private activities, bystanders, and patterns of attention. Potential applications include assistive visual interfaces, but gaze-conditioned systems could also enable intrusive monitoring. Access to gaze coordinates does not itself establish consent. Any deployment should obtain informed consent from affected users, limit data collection and retention, protect access to video and gaze records, and respect the source datasets' usage conditions. The reported benchmark experiments do not establish safety for deployment or energy savings on wearable hardware.

\subsection*{Reproducibility Statement}
Section~\ref{sec:method} describes the method, and Section~\ref{sec:exp-implementation} specifies the data split, metrics, and training settings. Section~\ref{sec:cross-dataset} describes transfer evaluation. Appendices~\ref{sec:detailed-ablation}--\ref{sec:gaze-robustness} specify the component comparisons and report across-seed variability for both GazeLens backbones. Appendix~\ref{sec:implementation-details} documents parameter counts and the FLOP measurement protocol.

\clearpage
\bibliographystyle{iclr2027_conference}
\bibliography{main}

\clearpage
\appendix
\raggedbottom
\setlength{\intextsep}{8pt}
\newcommand{\appendixtablefit}[1]{\begingroup\setbox0=\hbox{#1}\ifdim\wd0>\linewidth\resizebox{\linewidth}{!}{\usebox0}\else\usebox0\fi\endgroup}
\section{Additional Benchmark Results}
\label{sec:supp}

\subsection{Transfer Results by Source}

\label{sec:transfer-details}
This experiment tests whether the frame-selection benefit extends to a second benchmark, complementing the StreamGaze ablations. Following Section~\ref{sec:cross-dataset}, we train on the EGTEA source of EgoGazeVQA for four epochs and evaluate on held-out Ego4D and EgoExo examples. Table~\ref{tab:transfer-details} expands Figure~\ref{fig:transfer-efficiency} by source and compares uniform sampling with GazeLens's HOG-and-sharpness selector. The combined accuracy pools correct predictions across both sources.

\par\addvspace{\intextsep}
\noindent\begin{minipage}{\linewidth}
\makeatletter\def\@captype{table}\makeatother
\centering
\small
\caption{Frame-selection comparison on EgoGazeVQA. Each row reports a selector on the indicated evaluation set. \emph{Examples} is the number of evaluation items, and \emph{Frames} is the mean selected frames per item. Accuracy is in percent.}
\label{tab:transfer-details}
\begin{tabular}{llrrr}
\toprule
Evaluation set & Selector & Examples & Frames & Accuracy \\
\midrule
Combined & Uniform T16 & 1265 & 15.72 & 56.84 \\
Combined & Uniform T7 & 1265 & 7.00 & 58.10 \\
Combined & Uniform T8 & 1265 & 8.00 & 58.02 \\
Combined & GazeLens selector & 1265 & 5.74 & 57.71 \\
Ego4D & Uniform T16 & 577 & 15.90 & 51.30 \\
Ego4D & Uniform T7 & 577 & 7.00 & 51.99 \\
Ego4D & Uniform T8 & 577 & 8.00 & 51.82 \\
Ego4D & GazeLens selector & 577 & 5.83 & 52.17 \\
EgoExo & Uniform T16 & 688 & 15.58 & 61.48 \\
EgoExo & Uniform T7 & 688 & 7.00 & 63.23 \\
EgoExo & Uniform T8 & 688 & 8.00 & 63.23 \\
EgoExo & GazeLens selector & 688 & 5.67 & 62.35 \\
\bottomrule
\end{tabular}
\end{minipage}\par\addvspace{\intextsep}

GazeLens's selector exceeds Uniform T16 on both sources while retaining fewer than six frames per evaluation example on average. It also achieves the highest accuracy among the tested selectors on Ego4D, whereas T7 and T8 remain more accurate on EgoExo. Thus, the smaller frame budget transfers across both sources, while the accuracy trade-off depends on the source.

\section{Additional Ablations}
\label{sec:temporal-selection}
\label{sec:detailed-ablation}

We report across-seed variability for both backbones, then examine frame-selection cues and keyframe capacity on SmolVLM2-500M. Macro and Micro follow the definitions in Section~\ref{sec:exp-implementation}. Full-method means average three checkpoints.

\subsection{Across-Seed Variability}
To examine how training variability is distributed across tasks, Table~\ref{tab:seed-variability} reports task scores and aggregate accuracy across three seeds for each backbone. All checkpoints are evaluated on the same held-out examples.

\par\addvspace{\intextsep}
\noindent\begin{minipage}{\linewidth}
\makeatletter\def\@captype{table}\makeatother
\centering\small
\caption{GazeLens accuracy across three training seeds per backbone on the same held-out examples. Smaller values following the means denote sample SDs across checkpoints. These means supply the main comparison.}
\label{tab:seed-variability}
\appendixtablefit{\begin{tabular}{lccccc}
\toprule
Backbone & Past $\uparrow$ & Present $\uparrow$ & Proactive $\uparrow$ & Macro $\uparrow$ & Micro $\uparrow$ \\
\midrule
256M & 0.351\,{\scriptsize $\pm 0.053$} & 0.532\,{\scriptsize $\pm 0.010$} & 0.643\,{\scriptsize $\pm 0.006$} & 0.509 {\scriptsize $\pm 0.017$} & 0.617\,{\scriptsize $\pm 0.005$} \\
500M & 0.423\,{\scriptsize $\pm 0.041$} & 0.551\,{\scriptsize $\pm 0.011$} & 0.645\,{\scriptsize $\pm 0.011$} & 0.539\,{\scriptsize $\pm 0.015$} & 0.640\,{\scriptsize $\pm 0.005$} \\
\bottomrule
\end{tabular}}
\end{minipage}\par\addvspace{\intextsep}

Past has the largest across-seed variation for both backbones, whereas Micro has an SD of 0.005 in each case. Moving to the 500M backbone improves Past more than Present or Proactive. Consequently, its higher Macro primarily reflects improved Past accuracy.

\subsection{Selection Cues}
\label{sec:supp-hog-sharp}
HOG emphasizes changes between frames, while sharpness favors visually clear frames. Following the uniform-versus-dynamic comparison in Table~\ref{tab:ablate-uniform-dynamic}, Table~\ref{tab:ablate-hog-sharp} examines whether combining these cues improves the accuracy and frame-count trade-off over using either cue alone.

\par\addvspace{\intextsep}
\noindent\begin{minipage}{\linewidth}
\makeatletter\def\@captype{table}\makeatother
\centering\small
\caption{HOG vs.\ sharpness inside the dynamic picker. $\bar T$ is mean frames per evaluation example.}
\label{tab:ablate-hog-sharp}
\appendixtablefit{\begin{tabular}{lcc}
\toprule
Picker configuration & Macro & Mean $\bar{T}$ \\
\midrule
HOG only & 0.501 & 11.31 \\
Sharpness only & 0.508 & 13.23 \\
HOG + sharpness & 0.539\,{\scriptsize $\pm 0.015$} & 12.72 \\
\bottomrule
\end{tabular}}
\end{minipage}\par\addvspace{\intextsep}

The combined picker achieves the highest Macro even though it selects fewer frames than sharpness alone. This supports using the cues together to balance changes in scene content with image clarity within the frame budget.

\subsection{Keyframe Capacity}
\label{sec:supp-ablation}

To examine how much temporal capacity the selector needs, Table~\ref{tab:ablate-dynamic-cap} varies $t_{max}$ from 4 to 16 with $t_{min}=4$. The $t_{max}=16$ column reports the full-method configuration.

\par\addvspace{\intextsep}
\noindent\begin{minipage}{\linewidth}
\makeatletter\def\@captype{table}\makeatother
\centering
\caption{Impact of dynamic keyframe capacity ($t_{max}$) on StreamGaze macro and micro accuracy. A fixed $t_{min}=4$ is maintained.}
\label{tab:ablate-dynamic-cap}
\appendixtablefit{\begin{tabular}{l cccc}
\toprule
$t_{max}$ cap & 4 & 8 & 12 & 16 \\
\midrule
Macro & 0.505 & 0.526 & 0.536 & 0.539\,{\scriptsize $\pm 0.015$} \\
Micro & 0.612 & 0.629 & 0.637 & 0.640\,{\scriptsize $\pm 0.005$} \\
\bottomrule
\end{tabular}}
\end{minipage}\par\addvspace{\intextsep}

Both metrics improve as the cap increases, with the largest gains occurring between four and eight frames. Raising the cap from 12 to 16 adds 0.003 to both Macro and Micro. Thus, additional temporal capacity helps most at the smallest budgets, with diminishing accuracy gains as the cap grows. The full configuration allows up to 16 frames while selecting 12.72 per evaluation example on average.

\section{Spatial Representation}
\label{sec:additional-results}

These experiments compare pooling rules, token composition, and spatial token budgets on SmolVLM2-500M. Frame selection uses HOG and sharpness as described in Section~\ref{sec:method-dynamic-frame}. The full-method reference reports mean $\pm$ sample SD across three checkpoints. The comparisons below distinguish changes made at evaluation time from variants trained separately under the four-epoch protocol.

In these controls, $p_t$ denotes the primary context token and $m_t$ the global mean token, corresponding to $\mathbf{a}^{\text{ctx}}_t$ and $\mathbf{a}^{\text{vis}}_t$ in the method description.

\Needspace{18\baselineskip}
\subsection{Gaze-Weighted Pooling}
While Table~\ref{tab:ablate-gate-fixed} examines how the focused and complementary pools are mixed, Table~\ref{tab:ablate-gate-pool} tests whether the focused pool itself benefits from gaze weighting. It replaces gaze-weighted pooling in the primary branch with the spatial mean while retaining complementary pooling and the global token.

\par\addvspace{\intextsep}
\noindent\begin{minipage}{\linewidth}
\makeatletter\def\@captype{table}\makeatother
\centering\small
\caption{Gaze-weighted pooling vs.\ average pooling. \emph{Mean pool} sets $\mathbf{g}_t$ to the spatial mean of $F_t$ before gating with $\mathbf{c}_t$.}
\label{tab:ablate-gate-pool}
\appendixtablefit{\begin{tabular}{lc}
\toprule
Primary branch pooling & Macro \\
\midrule
Gaze-weighted pool & 0.539\,{\scriptsize $\pm 0.015$} \\
Spatial mean pool & 0.520 \\
\bottomrule
\end{tabular}}
\end{minipage}\par\addvspace{\intextsep}

Replacing gaze weighting with a spatial mean lowers Macro by 0.019. This comparison supports preserving a gaze-focused summary before the gate blends it with complementary context, even when a separate global token is available.

\subsection{Representation and Spatial Budget}
To assess which information deserves a limited token budget, Table~\ref{tab:new-representations} compares token representations and spatial budgets. Primary-only, global-only, and duplicated-primary rows modify the reference checkpoint at evaluation time, whereas the focus-plus-complement and spatial-grid variants use separate training runs. Each row specifies the tokens supplied to the language model per selected frame.

\par\addvspace{\intextsep}
\noindent\begin{minipage}{\linewidth}
\makeatletter\def\@captype{table}\makeatother
\centering\small
\caption{StreamGaze representation and spatial-budget controls on the full test set. Tokens are per frame, and accuracies are percentages. Ours reports three-seed mean $\pm$ SD.}
\label{tab:new-representations}
\appendixtablefit{\begin{tabular}{lrrr}
\toprule
Representation & Tokens/frame & Macro & Micro \\
\midrule
Primary $[p_t]$ & 1 & 52.0 & 62.5 \\
Global $[m_t]$ & 1 & 49.5 & 60.5 \\
Focus + complement $[g_t,c_t]$ & 2 & 52.5 & 63.0 \\
Primary pair $[p_t,p_t]$ & 2 & 52.1 & 62.6 \\
Ours $[p_t,m_t]$ & 2 & 53.94\,{\scriptsize $\pm 1.54$} & 64.01\,{\scriptsize $\pm 0.49$} \\
$2\times2$ grid & 4 & 54.0 & 64.2 \\
$4\times4$ grid & 16 & 54.3 & 64.5 \\
\bottomrule
\end{tabular}}
\end{minipage}\par\addvspace{\intextsep}

Pairing the primary and global tokens gives the highest Macro and Micro among the one- and two-token variants, while duplicating the primary token brings little gain. The $4\times4$ grid improves Macro by 0.36 percentage points and Micro by 0.49 percentage points over our pair, but uses eight times as many tokens per frame. These comparisons support allocating the two-token budget to a question-conditioned primary summary and a global scene summary.

\Needspace{24\baselineskip}
\section{Gaze Robustness}
\label{sec:gaze-robustness}

To assess the gaze input, we evaluate sensitivity to coordinate noise and quantify fixation coverage. Macro and Micro use the same definitions as the main comparison. The zero-noise reference reports mean $\pm$ sample SD across three checkpoints.

\subsection{Robustness to Gaze Noise}
\label{sec:supp-robustness}

To test sensitivity to imprecise gaze, Table~\ref{tab:ablate-gaze-noise} varies Gaussian noise in normalized gaze coordinates at inference time. The zero-noise column reports the full-method result.

\par\addvspace{\intextsep}
\noindent\begin{minipage}{\linewidth}
\makeatletter\def\@captype{table}\makeatother
\centering
\caption{Model performance under synthetic spatial gaze noise at test time.}
\label{tab:ablate-gaze-noise}
\begin{tabular}{l cccccc}
\toprule
Noise $\sigma$ & 0.00 & 0.01 & 0.02 & 0.05 & 0.10 & 0.20 \\
\midrule
Macro & 0.539\,{\scriptsize $\pm 0.015$} & 0.537 & 0.534 & 0.526 & 0.514 & 0.498 \\
Micro & 0.640\,{\scriptsize $\pm 0.005$} & 0.638 & 0.635 & 0.627 & 0.615 & 0.599 \\
\bottomrule
\end{tabular}
\end{minipage}\par\addvspace{\intextsep}

At $\sigma=0.02$, both Macro and Micro are 0.005 below the zero-noise reference. At the largest tested noise level of $\sigma=0.20$, both decrease by 0.041. Accuracy therefore changes little under small perturbations and declines as the noise grows, indicating that the compact representation tolerates modest coordinate errors but remains sensitive to gaze quality.

\subsection{Fixation Coverage}
Fixation coverage is measured on a separate cached selection trace containing 17,950 selected-frame uses. Of these, 82.84\% fall inside detected fixation intervals, 7.28\% have valid coordinates outside those intervals, and 9.88\% lack valid coordinates. Repeated uses of a frame across questions or probes are counted separately. These percentages describe frame coverage by fixation status and coordinate availability.

\section{Implementation and Computation}
\label{sec:implementation-details}

\subsection{Trainable Parameters}
\label{sec:supp-paramdetail}

Our method trains 1.75M parameters on the 256M backbone and 3.88M on the 500M backbone, including LoRA. Gaussian pooling and the frozen question embedding introduce no new parameters. Linear layers use biases, and LoRA uses rank $r=8$ and $\alpha=16$ on $q_{\text{proj}}$ and $v_{\text{proj}}$ in the vision encoder and language model.

\subsection{Controlled inference cost}
\label{sec:controlled-efficiency}
We count input-forward computation at batch size one on an NVIDIA RTX 3080, using synthetic frames and a fixed question-and-options prompt. The native processed frame sizes are $512\times512$ for SmolVLM2, $1024\times1024$ for FastVLM, and $224\times224$ for both Mobile-VideoGPT vision encoders. SmolVLM2 and FastVLM use bfloat16, whereas Mobile-VideoGPT uses float16. Full-grid backbones are pretrained, and adapted SmolVLM2 methods add randomly initialized rank-eight LoRA adapters and method modules. Parameter values do not change the counted tensor shapes.

FLOPs cover supported operators in the vision encoders, token modules, and language model, including all prompt-position logits. A multiply-add counts as two operations, and grouped-query attention uses dense-equivalent attention arithmetic, including the full causal matrix convention. These are arithmetic counts, not hardware instruction counts. Video decoding, external frame selection, preprocessing, model loading, and autoregressive answer generation are excluded. Normalization and nonlinear operations are not included in the operator-counting convention.

Baselines use 16 frames. For both adaptive backbones, we count each frame length from 4 to 16 and weight it by the 2,418-example test-set selector distribution, averaging $12.72$ frames per example. Per-example frame counts are identical across all three runs of both the 256M and 500M variants. The adaptive costs combine temporal selection and spatial compression.

Mobile-VideoGPT additionally uses an opaque selective-scan kernel. From the observed tensor dimensions, we estimate its arithmetic as $BLD(7N+2)$ per directional scan, where $B$ is batch size, $L$ sequence length, $D$ inner width, and $N$ state size. This counts state updates, output contraction, and the skip connection, excluding exponentials and softplus. Its 64 observed scans add $0.013187$ TFLOPs to $3.889336$ operator-counted TFLOPs. The total combines operator counts with this analytical scan estimate.

\section{Limitations and Future Work}
\label{sec:limitations}

GazeLens targets devices with gaze sensors. When coordinates are missing, pooling becomes approximately uniform and loses the spatial prior that distinguishes attended content. A useful next step is to estimate gaze confidence and adapt the pooling rule when tracking is unreliable, so that uncertain fixation measurements do not determine the spatial representation.

Temporal selection uses HOG transitions and image sharpness without conditioning on the question. Consequently, these cues can suppress visually repetitive frames that still contain relevant evidence. Future work could incorporate question relevance into selection while keeping its cost below that of encoding discarded frames. Selected frames are currently encoded independently, so reusing visual tokens across unchanged regions is another route to reducing repeated computation during longer interactions. Such extensions should be evaluated with end-to-end latency and memory measurements, since the reported model-forward FLOPs exclude frame selection, decoding, and preprocessing.

Finally, two tokens per frame provide a compact input but can discard spatial detail needed for fine-grained reasoning. The spatial-budget comparisons motivate adapting token capacity to the question and scene complexity. Increasing capacity only when the task needs finer evidence could extend the present accuracy--computation trade-off, while retaining a compact representation for simpler inputs.

\end{document}